# Using natural language processing and structured medical data to phenotype patients hospitalized due to COVID-19

Author: Feier Chang[*1], Jay Krishnan[*2], Jillian H Hurst[3], Michael E Yarrington[2], Deverick J Anderson[2], Emily C O'Brien[4,5], Benjamin A Goldstein[1,3-5]

1. Department of Biostatistics & Bioinformatics, Duke University, Durham, NC
2. Department of Medicine, Duke University, Durham, NC
3. Department of Pediatrics, Duke University, Durham, NC
4. Department of Population Health Sciences, Duke University, Durham, NC
5. Duke Clinical Research Institute, Duke University, Durham, NC

* Indicates first co-authors.

Corresponding author: Benjamin A Goldstein

Department of Biostatistics & Bioinformatics, Duke University, 2424 Erwin Road, 9023, Durham, NC, 27705

ben.goldstein@duke.edu (919)681-5011

**Key Points**

**Question:**

Can we classify hospitalized patients who test positive for COVID-19 as being hospitalized due to COVID-19 versus hospitalized for other reasons?

**Findings:**

A classification algorithm that incorporated provider notes performs significantly better than an algorithm using only structured electronic health records data elements; such classifications can impact subsequent association analyses.

**Meaning:**

Tools are needed to accurately identify cause-specific outcomes such as hospitalization due to COVID-19 in order to conduct accurate assessments of related association analyses.

**Abstract**

**Importance:**

Throughout the COVID-19 pandemic, many hospitals conducted routine testing of hospitalized patients for SARS-CoV-2 infection upon admission. Some of these patients are admitted for reasons unrelated to COVID-19, and incidentally test positive for the virus. Because COVID-19-related hospitalizations have become a critical public health indicator, it is important to identify patients who are hospitalized because of COVID-19 as opposed to those who were admitted for other indications.

**Objective:**

We compared the performance of different computable phenotype definitions for COVID-19 hospitalizations that use different types of data from the electronic health records (EHR), including structured EHR data elements, provider notes, or a combination of both data types.

**Design:**

Retrospective data analysis utilizing chart review-based validation.

**Setting:**

Large academic medical center

**Participants:**

586 hospitalized individuals who tested positive for SARS-CoV-2 during January 2022.

**Exposure:**

Not applicable.

**Main Outcomes and Measures:**

We used LASSO regression and Random Forests to fit classification algorithms that incorporated structured EHR data elements, provider notes, or a combination of structured data and provider notes. We used natural language processing to incorporate data from provider notes. The performance of each model was evaluated based on Area Under the Receiver Operator Characteristic (AUROC) and an associated decision rule based on sensitivity and positive predictive value. We also identified top words and clinical indicators of COVID-19-specific hospitalization and assessed the impact of different phenotyping strategies on estimated hospital outcome metrics.

**Results:**

Based on a chart review, 38% of 586 patients were determined to be hospitalized for reasons other than COVID-19 despite having tested positive for SARS-CoV-2. A classification algorithm that used provider notes had significantly better discrimination than one that used structured EHR data elements (AUROC: 0.894 vs 0.841, p < 0.001), and performed similarly to a model that combined provider notes with structured data elements (AUROC: 0.894 vs 0.893). Assessments of hospital outcome metrics significantly differed based on whether the population included all hospitalized patients who tested positive for SARS-CoV-2 versus those who were determined to have been hospitalized due to COVID-19.

**Conclusion and Relevance:**

These findings highlight the importance of cause-specific phenotyping for COVID-19 hospitalizations. More generally, this work demonstrates the utility of natural language processing approaches to derive information related to patient hospitalizations in cases where there may be multiple conditions that could serve as the primary indication for hospitalization.

**INTRODUCTION**

Hospitalization due to coronavirus disease 2019 (COVID-19) has become a key public health indicator. One of the primary goals of vaccination against SARS-CoV-2, the etiological agent of COVID-19, is to reduce the incidence of severe disease and death, with hospitalization serving as a primary endpoint in vaccine efficacy trials [1]. Further, hospitalization has become a primary indicator of community transmission levels of SARS-CoV-2 infection [2], including disease severity and health system capacity [3-6]. Similarly, hospitalization due to COVID-19 is a typical outcome of interest in public health studies of COVID-19 using real-world data sources such as electronic health records (EHR) data [7-10]. Finally, because of the rise of rapid, at-home testing for SARS-CoV-2 infection, COVID-19 cases that do not rise to the level of requiring medical attention are likely to be missed or underreported, affecting assessments of COVID-19 prevalence [11]. Thus, there is a critical need to rapidly and accurately identify hospitalizations due to COVID-19.

Due to concerns related to hospital-based spread of SARS-CoV-2, many institutions routinely perform SARS-CoV-2 testing in patients admitted to the hospital, regardless of the primary reason for admission [12,13]. While SARS-CoV-2 testing is important to guide care and ensure that providers take precautions to prevent infection, such routine testing potentially complicates retrospective studies using real-world data sources. Specifically, it becomes challenging to distinguish a patient admitted because of COVID-19 from a patient who incidentally tested positive for SARS-CoV-2 infection. In both cases, patients would have a positive lab test and would (presumably) have an ICD-10 code for COVID-19. Previous reports have noted that incidental positives may account for around 26% of all COVID-19 positive patients [14].

Given the public health importance of identifying hospitalizations due to COVID-19 rather than hospitalizations where SARS-CoV-2 infection was identified incidentally, methods, i.e., computable phenotypes, are needed to distinguish the two conditions in retrospective data sources. Such phenotypes would be instrumental in retrospective studies of COVID-19 patients and in public health surveillance. To address this need, we explore the potential to use both structured data (i.e., diagnosis codes, medications, procedure codes) and unstructured data (i.e., provider notes) to identify patients admitted because of COVID-19 versus patients who incidentally tested positive for SARS-CoV-2 during admission.

## MATERIALS AND METHODS

**Study Setting**

We performed a retrospective study of patients $\geq$18 years of age who were hospitalized with a documented positive SARS-CoV-2 test during January 2022. We conducted our study at Duke University Health System (DUHS), which consists of a quaternary academic medical center and two associated community-based hospitals. This study was designated as exempt human subjects research by the DUHS IRB.

**Study Data**

Using DUHS EHR data, we identified all patients admitted during the week of 01/16/22-01/22/22 with documentation of a positive SARS-CoV-2 test in the prior 20 days based on infection isolation status, a clinical EHR tool used to indicate whether COVID-19-specific personal protective equipment is required when entering a patient's room. Charts from this week were specifically reviewed in part due to a data request from the North Carolina Division of Public Health to understand the epidemiology of COVID-19-related hospitalizations. We excluded individuals with resolved COVID-19 isolation status as well as those admitted prior to 01/01/22 to create a cohort of patients who were likely to be infected with the Omicron variant of SARS-CoV-2. During this period, Omicron was the predominant SARS-CoV-2 variant in circulation in the United States and was associated with the largest wave [8] of SARS-CoV-2 infections to date. For each patient, we extracted medical record number, date of admission, hospital unit, and level of care.

To generate a criterion standard for classification, six trained healthcare providers manually reviewed patient records for the index admission to adjudicate whether each patient's SARS-CoV-2 infection was the primary reason for admission or an incidental finding. Providers attributed hospitalizations as due to COVID-19 if admissions were due to primary manifestations of SARS-CoV-2 infection, such as hypoxia or need for supplemental oxygen, or COVID-19-associated complications, such as dehydration or weakness.

*Analytic Data*

For each admission reviewed, we extracted structured EHR data elements from their hospitalization from the Duke Clinical Research Datamart, an EHR database that is based on an extension of the PCORnet Common Data Model [15]. Provider notes were extracted from Duke's Electronic Data Warehouse. Extracted structured data elements included demographics, service encounter characteristics, diagnoses, laboratory tests, COVID-19 vaccination status, and medications (see **Supplemental Table 1**). Provider notes included emergency department admission notes, progress notes, operative notes, history and physical examination notes, and discharge summaries.

*Provider Note Analysis*

To analyze the provider notes, we used the term frequency-inverse document frequency (TF-IDF) approach. TF-IDF [16] generates a numeric value for each word within a set of notes for a given patient. The word value is based on how common the word is in a patient's set of notes (TF), divided by how common the word is across all the patient's notes (IDF), resulting in a numeric representation for each word on a per-patient basis. While this is a simple word-based representation, there are two advantages to this approach compared to deep learning embedding-based approaches: 1) it is possible to directly assess the importance of individual words, and 2) TF-IDF tends to be more robust with small data sets. We used the nltk package in Python [17] to tokenize words into a dictionary, calculate word counts, and generate the corresponding weight matrix. We removed any words that appeared fewer than 50 times.

**Analytic Approach**

We first described the clinical characteristics of patients hospitalized due to COVID-19 versus those with incidental COVID-19 using standardized mean differences (SMD), with an SMD of 0.10 indicative of a clinical meaningful difference. Next, we developed three classification models for COVID-19-specific hospitalization, one based entirely on features derived from structured EHR data elements, a second based on provider notes alone, and a third using both structured data elements and provider notes. We used LASSO logistic regression and Random Forests to estimate the model. Due to the relatively small sample size, we present our results based on 10-fold cross-validation. We performed TF-IDF separately within each cross-validation fold.

We evaluated the six classification models by calculating the area under the receiver operator characteristic curve (AUROC) along with associated 95% confidence intervals. We identified the top clinical features and words that appeared in provider notes based on the LASSO and Random Forests models. We plotted the precision-recall curve to better understand the performance of a classification model and assessed the impact of different-rule based phenotypes.

To understand the potential impact of accurate phenotyping, we performed an illustrative association analysis evaluating the relationship between vaccination status and different hospital outcome metrics: length of stay, intensive care unit (ICU) utilization, and in-hospital mortality. These were chosen since they are standard quality metric for operational purposes. We regressed each outcome onto vaccination status. We used a log-linear model for length of stay and logistic regression for ICU utilization and in-hospital mortality. We performed each regression using the full cohort and compared it to a model using a dataset using only the adjudicated COVID-19 hospitalized encounters. We also tested for an interaction between vaccination status and cause of hospitalization. We emphasize that these are illustrative analyses and are not meant to infer any causal effects of vaccination, but rather serve to illustrate the importance of using cause-specific phenotyping for relevant COVID-19 outcomes.

All work was performed in R version 4.1.2 [18] and Python version 3.9.1 [19].

## RESULTS

In total, we reviewed the charts of 630 patients who were admitted and tested positive for SARS-CoV-2. After excluding patients less than 18 years-old and with privacy restrictions, our dataset included 586 unique patients who were hospitalized and had tested positive for SARS-CoV-2. Of these, 38% were determined through clinician review to be hospitalized for reasons other than COVID-19. During their assessments, our chart reviewers noted that it was often readily apparent which hospitalizations were attributable to COVID-19 and which were not.

Characteristics by admission cause are shown in **Table 1**. Compared with patients hospitalized for indications other than COVID-19, patients hospitalized due to COVID-19 were on average older (mean age 62.7 vs. 51.9 years, SMD 0.587) and their admissions were more commonly labeled as emergency admission (95.6% vs. 73.7%, SMD 0.641). Furthermore, patients hospitalized due to COVID-19 were substantially more likely to receive COVID-19 therapies including steroids (64.4% vs. 24.1%, SMD 0.887) and the anti-viral agent remdesivir (68.2% vs. 24.6%, SMD 0.974) during their hospitalization. Patients hospitalized due to COVID-19 had lower lymphocyte counts on average compared with those hospitalized for reasons other than COVID-19. Normal levels of C-reactive protein and the lack of d-dimer testing were associated with hospitalizations for reasons other than COVID-19.

### Performance of Classification models

After tokenizing words and removing terms with fewer than 50 occurrences, our models included 7953 unique terms. There was minimal difference between the LASSO and Random Forests models (**Table 2**). The model based solely on provider notes (AUROC: 0.894 95% CI: 0.868, 0.920) had better discrimination than the model based solely on structured data elements (AUROC: 0.841 95% CI: 0.809, 0.874) (p < 0.001). The model using both provider notes and structured data elements (AUROC: 0.892 95% CI: 0.868, 0.919) had similar discrimination to the model based solely on provider notes (p = 0.909).

Next, we examined the top structured data elements and terms in each model (Figure 2a). Highly predictive data elements and words corresponded to patient characteristics with large SMDs (**Table 1**). Predictive words reflected clinical care for COVID-19 (in the positive direction) as compared to other reasons for care (negative direction). Terms reflective of COVID-19-specific hospitalization were related to the care of COVID-19 patients, such as "remdesivir" and "dexamethason". Other structured elements related to the likelihood of being hospitalized for COVID-19 included receipt of steroids, low lymphocyte counts, and underweight BMI. Terms reflective of hospitalizations due to indications other than COVID-19 included strings that may be related to surgical procedures (such as "surgic" for "surgical" or "dress" for "dressing"). For structured data elements, a lack of d-dimer collection and low ferritin were most commonly associated with admissions for reasons other than COVID-19. Similar features were identified from the Random Forests model (**Supplemental Figure 1**).

**Impact of Correct Classification**

In order to assess the performance of a computable phenotype-based decision rule, we examined the precision-recall curve of the different models (Figure 1). For example, a rule that maintained a sensitivity of 90% (i.e., would capture 90% of all patients truly hospitalized due to COVID) results in a positive predictive value of 76%, 82% and 84% based on structured data elements, provider notes, and their combination, respectively. To illustrate the impact of these differences, we consider the impact of implementing each of these phenotypes at a 90% sensitivity to classify patients during the January Omicron wave. Within our health system, 1378 people were hospitalized and tested positive for SARS-CoV-2. Based on our analyses utilizing the LASSO based phenotype incorporating structured data, provider notes, or their combination, would result in approximately 244, 165, and 142 false positives, respectively.

Finally, we assessed the impact of accurate phenotyping on our understanding of different hospital outcome metrics. **Table 3** presents regression results for the marginal relationship between vaccine status and three outcomes across the two analytic cohorts, wherein unvaccinated patients serve as the reference group. For length of stay, the magnitude of the effect of vaccine status changes based on the cohort used. In the cohort of all hospitalized patients, vaccinated patients had a shorter length of stay (Relative Rate: 0.81 95% CI: 0.71, 0.93). However, when limiting the analytic cohort to patients hospitalized due to COVID-19, there is no significant difference in length of stay for vaccinated versus unvaccinated patients (RR: 0.98 95% CI: 0.83, 1.16; p-value for interaction <0.001). We found similar patterns in analyses of other in-hospital outcomes, with vaccination associated with reduced risk of ICU utilization and in-hospital mortality amongst those hospitalized for reasons other than COVID-

19 than among those hospitalized due to COVID-19. Effects were robust to adjustment for age (**Supplemental Table 2**).

## DISCUSSION

Due to the public health importance of accurate identification of COVID-19-related hospitalizations, there is a need for methods and computable phenotypes to identify hospital admissions in which the primary cause is COVID-19 [20]. We used machine learning methods and physician chart review to develop a classification algorithm for hospitalization due to COVID-19. We found that 38% of patients hospitalized at our institution during the Omicron wave and who tested positive for SARS-CoV-2 infection were hospitalized for reasons other than COVID-19. These findings are in line with other recent studies, which found that an average of 26% of hospitalized patients with a positive SARS-CoV-2 test had a primary indication for hospitalization unrelated to COVID-19 [14]. We found that a model based on provider notes performed better than one based solely on structured EHR data elements. This work has important implications for retrospective analyses using EHR data to assess outcomes related to COVID-19, including vaccine effectiveness and health system capacity [21].

Prior work by Lynch and colleagues has evaluated the utility of ICD-10 codes for COVID-19 diagnosis in inpatient, outpatient, and emergency/urgent care settings during time periods across the pandemic. Using a weighted, random sample of 1500 records from the Department of Veterans Affairs (VA), they found that the COVID-19 ICD-10 code (U07.1) had a relatively low positive predictive value that across settings and time periods [22]. These findings highlight the need for additional contextual data to identify acute cases of COVID-19. The Consortium for Clinical Characterization of COVID-19 by EHR (4CE) conducted a similar study of EHR data from 12 clinical sites to identify combinations of structured data elements to generate a reliable computable phenotype for hospitalization due to COVID-19, with a reported AUROC of 0.903 [23]. Similarly, we derived an AUROC of 0.841 based solely on structured data elements; however, we also found that inclusion of provider notes significantly improved the performance of the classification model (AUROC of 0.893). This result is not surprising as the clinical narrative often includes important nuance, and as our chart reviewers noted, it was often readily apparent which hospitalizations were attributable to COVID-19 and which were not. Of note, chart reviewers in our study classified hospitalizations indirectly due to SARS-CoV-2 infection, such as COVID-19-related weakness or delirium, as hospitalizations due to COVID-19, which could partly explain the observed discriminatory ability between our study and the study conducted by 4CE.

By using the TF-IDF approach in conjunction with LASSO regression, we identified both individual terms and the direction of the association between each term and the hospitalization indication. While TF-IDF is a simple natural language process approach, it is also very scalable, interpretable, and implementable. These results highlight the power of even simple natural language models. The terms that best predicted hospitalizations due to COVID-19 included common descriptors used in the clinical care of COVID-19 patients, such as "hypox" (likely shortened from "hypoxia" or "hypoxic") or COVID-19 therapies like remdesivir. Conversely, the terms not associated with hospitalizations due to COVID-19 included words related to surgery, a common indication for hospital admission generally unrelated to COVID-19 infection.

We also assessed the real-world impact of using a phenotype for COVID-19 specific hospitalization. In studying hospitalized patients with COVID-19, the simplest analysis would be to include all patients with a COVID-19 positive test. As our illustrative analysis showed, using this full – but heterogeneous – cohort, would suggest that vaccination status is associated with shorter length of stay. However, if we reduce the analysis to only include patients identified as having been hospitalized due to COVID-19, i.e., people with symptoms of COVID-19, the analysis indicates that vaccines are not associated with shorter length of stay. We interpret these data as indicating that conditional on someone being sick enough to be hospitalized due to COVID-19, vaccines provide no additional benefit in terms of the length of hospitalization. Similar patterns were found for other hospital outcome metrics. While this analysis is not intended to be a causal analysis, it does illustrate how the use of accurately classified cohorts is important for the calculation of standard outcome metrics and would likely impact other related association analyses.

More broadly, this work highlights the importance, and challenge, of phenotyping cause specific events. While there is a rich literature on computable phenotypes, most of these are geared towards identification of chronic disease (e.g., presence of asthma). However, few computable phenotypes have focused on cause-specific events (i.e., asthma exacerbation). Such cause specific phenotypes often suffer from poor specificity and can require algorithms that are more complex than those required for chronic conditions. As this work shows, and suggested by others, NLP-based phenotyping approaches are becoming more common, and further comparisons between NLP approaches and other methods will be needed to determine whether using text data can improve cause-specific phenotypes.

While our study used rigorous methods, there are some key limitations. Most importantly, this study was conducted across a single hospital system and may not be reflective of practices at other institutions. Importantly, we would not expect our specific phenotype algorithm to be generalizable to other institutions. Second, we only looked at one period of time, namely the January 2022 Omicron wave; however, there are documented differences in in the rate of hospitalization and positive tests over the course of the pandemic and these models may not accurately reflect distinguishing factors during other waves. Finally, given the time constraints of chart-review we were only able to analyze a relatively small sample.

**CONCLUSIONS**

Overall, our results show that a sizable number of people who were hospitalized and tested positive for SARS-CoV-2 were hospitalized for reasons other than COVID-19. Conflation of these individuals can impact our understanding of hospital outcome metrics. We constructed a strong classification model that can be used as a computable phenotype to distinguish patients hospitalized due to COVID-19 versus those who incidentally tested positive for SARS-CoV-2 but were hospitalized for other reasons. Moreover, we found that while structured data elements are useful in constructing such a phenotype, provider notes had a higher positive predictive value than structured data elements alone. Future work should seek to explore the generalizability of such phenotypes across institutions as well as different waves of the COVID-19 pandemic.

**Table1: Cohort Description**

| Hospitalized due to COVID-19 | No | Yes | Total | Standardized Mean Difference |
|---|---|---|---|---|
|  | (N=224) | (N=362) | (N=586) |  |
| Sex (Female) | 120(53.6%) | 181(50.0%) | 301(51.4%) | 0.072 |
| Age (years) |  |  |  | 0.587 |
| Mean | 51.9 | 62.7 | 58.6 |  |
| Patient outcome at discharge |  |  |  | 0.169 |
| Dead | 18(8.0%) | 39(10.8%) | 57(9.7%) |  |
| Home | 176(78.6%) | 258(71.3%) | 434(74.1%) |  |
| Other Facility | 30(13.4%) | 65(18.0%) | 95(16.2%) |  |
| Admission Type |  |  |  | 0.641 |
| Emergency Admission | 165(73.7%) | 346(95.6%) | 511(87.2%) |  |
| Routine Elective Admission | 24(10.7%) | 4(1.1%) | 28(4.8%) |  |
| Urgent Admission | 35(15.6%) | 12(3.3%) | 47(8.0%) |  |
| Transfer to ICU | 45(20.1%) | 78(21.5%) | 123(21.0%) | 0.036 |
| Encounter Type |  |  |  | 0.181 |
| Emergency | 2(0.9%) | 1(0.3%) | 3(0.5%) |  |
| Emergency to Inpatient | 180(80.4%) | 314(86.7%) | 494(84.3%) |  |
| Inpatient | 31(13.8%) | 35(9.7%) | 66(11.3%) |  |
| Observation Stay | 11(4.9%) | 12(3.3%) | 23(3.9%) |  |
| Race/Ethnicity |  |  |  | 0.168 |
| Hispanic | 21(9.4%) | 20(5.5%) | 41(7.0%) |  |
| Non-Hispanic Black | 106(47.3%) | 175(48.3%) | 281(48.0%0 |  |
| Non-Hispanic White | 90(40.2%) | 152(42.0%) | 242(41.3%) | |
| Non-Hispanic Asian | 7(3.1%) | 14(3.9%) | 21(3.6%) | |
| Other races | 0(0%) | 1(0.3%) | 1(0.2%) | |
| Length of Stay | | | | 0.026 |
| Mean | 10.2 | 9.9 | 10 | |
| BMI | | | | 0.203 |
| Missing | 9(4%) | 9(2.5%) | 18(3.1%) | |
| Normal | 65(29.0%) | 89(24.6%) | 154(26.3%) | |
| Obese | 85(37.9%) | 147(40.6%) | 232(39.6%) | |
| Overweight | 60(26.8%) | 98(27.1%) | 158(27.0%) | |
| Underweight | 5(2.2%) | 19(5.2%) | 24(4.1%) | |
| Raw Payer Type Value | | | | 0.305 |
| Private | 102(45.5%) | 180(49.7%) | 282(48.1%) | |
| Public | 88(39.3%) | 144(39.8%) | 232(39.6%) | |
| Self-Pay | 21(9.4%) | 9(2.5%) | 30(5.1%) | |
| Other | 13(5.8%) | 29(8.0%) | 42(7.2%) | |
| Vaccinated against COVID-19 | 113(50.4%) | 178(49.2%) | 291(49.7%) | 0.026 |
| Comorbidities | | | | |
| Surgery | 200(89.3%) | 302(83.4%) | 502(85.7%) | 0.171 |
| Cancer | 29(12.9%) | 45(12.4%) | 74(12.6%) | 0.015 |
| Cardiovascular | 75(33.5%) | 146(40.3%) | 221(37.7%) | 0.142 |
| Hypertension | 73(32.6%) | 151(41.7%) | 224(38.2%) | 0.19 |
| Chronic Liver Disease | 30(13.4%) | 46(12.7%) | 76(13.0%) | 0.02 |
| Chronic Obstructive Pulmonary Disease | 21(9.4%) | 50(13.8%) | 71(12.1%) | 0.139 |
| Asthma | 18(8.0%) | 39(10.8%) | 57(9.7%) | 0.094 |
| Chronic Renal Disease | 44(19.6%) | 111(30.7%) | 155(26.5%) | 0.256 |
| Diabetes | 45(20.1%) | 103(28.5%) | 148(25.3%) | 0.196 |
| Medications | | | | |
| Bronchodilator | 44(19.6%) | 159(41.2%) | 193(32.9%) | 0.481 |
| Steroid | 54(24.1%) | 233(64.4%) | 287(49.0%) | 0.887 |
| Anticoagulant Antiplatelet | 121(54.0%) | 284(78.5%) | 405(69.1%) | 0.535 |
| Diuretic | 60(26.8%) | 131(36.2%) | 191(32.6%) | 0.203 |
| Cough Suppressant | 44(19.6%) | 162(44.8%) | 206(35.2%) | 0.558 |
| Paralytic | 10(4.5%) | 30(8.3%) | 40(6.8%) | 0.157 |
| Expectorant | 14(6.3%) | 56(15.5%) | 70(11.9%) | 0.3 |
| Remdesivir | 55(24.6%) | 247(68.2%) | 302(51.5%) | 0.974 |
| Inhaled Steroid | 24(10.7%) | 42(11.6%) | 66(11.3%) | 0.028 |
| Laboratory Tests | | | | |
| Lymphocyte Count Absolute | | | | 0.345 |
| High | 1(0.4%) | 2(0.6%) | 3(0.5%) | |
| Low | 12(5.4%) | 47(13.0%) | 59(10.1%) | |
| Normal | 23(10.3%) | 59(16.3%) | 82(14.0%) | |
| Not Taken | 188(83.9%) | 254(70.2%) | 442(75.4%) | |
| Lymphocyte Count | | | | 0.528 |
| Low | 17(7.6%) | 71(19.6%) | 88(15.0%) | |
| Normal | 131(58.5%) | 233(64.4%) | 364(62.1%) | |
| Not Taken | 76(33.9%) | 56(15.5%) | 132(22.5%) | |
| High | 0(0%) | 2(0.6%) | 2(0.3%) | |
| C-reactive Protein | | | | 0.602 |
| High | 62(27.7%) | 203(56.1%) | 265(45.2%) | |
| Normal | 11(4.9%) | 9(2.5%) | 20(3.4%) | |
| Not Taken | 151(67.4%) | 150(41.4%) | 301(51.4%) | |
| Ferritin | | | | 0.361 |
| High | 39(17.4%) | 107(29.6%) | 146(24.9%) | |
| Low | 2(0.9%) | 3(0.8%) | 5(0.9%) | |
| Normal | 17(7.6%) | 44(12.2%) | 61(10.4%) | |
| Not Taken | 166(74.1%) | 208(57.5%) | 374(63.8%) | |
| D-Dimer | | | | 1.187 |
| High | 19(8.5%) | 117(32.3%) | 136(23.2%) | |
| Normal | 36(16.1%) | 156(43.1%) | 192(32.8%) | |
| Not Taken | 169(75.4%) | 89(24.6%) | 258(44.0%) | |
| Procalcitonin | | | | 0.524 |
| High | 4(1.8%) | 22(6.1%) | 26(4.4%) | |
| Missing | 208(92.9%) | 268(74.0%) | 476(81.2%) | |
| Normal | 12(5.4%) | 72(19.9%) | 84(14.3%) | |

Table 2: Area Under the Receiver Operator Characteristic (AUROC) for Phenotype Classification Models

| Method/Category | Structured EHR Data elements | Provider notes | Structured Data + Notes |
|---|---|---|---|
| Lasso | 0.841(0.809-0.874) | 0.894(0.868-0.920) | 0.893(0.868-0.919) |
| Random Forests | 0.829(0.794-0.864) | 0.882(0.855-0.909) | 0.890(0.864-0.916) |

| Table 3: Marginal association between vaccine status* and outcome metrics, unadjusted for age **Outcome** | Full Cohort (95% CI) | Hospitalized Due to COVID-19 (95% CI) | Hospitalization unrelated to COVID-19 (95% CI) | P-value Hospitalization due to COVID-19 vs. Hospitalization unrelated to COVID-19 |
|---|---|---|---|---|
| Length of Stay (Relative Rate) | 0.81 (0.71, 0.93) | 0.98 (0.83, 1.16) | 0.59 (0.47, 0.74) | 0.0003 |
| ICU Utilization (Odds Ratio) | 1.04 (0.70, 1.56) | 1.25 (0.75, 2.07) | 0.77 (0.40, 1.49) | 0.26 |
| Mortality (Odds Ratio) | 1.02 (0.59, 1.78) | 1.45 (0.74, 2.88) | 0.48 (0.16, 1.29) | 0.08 |

* Unvaccinated as referenced group

Figure1: AUPRC

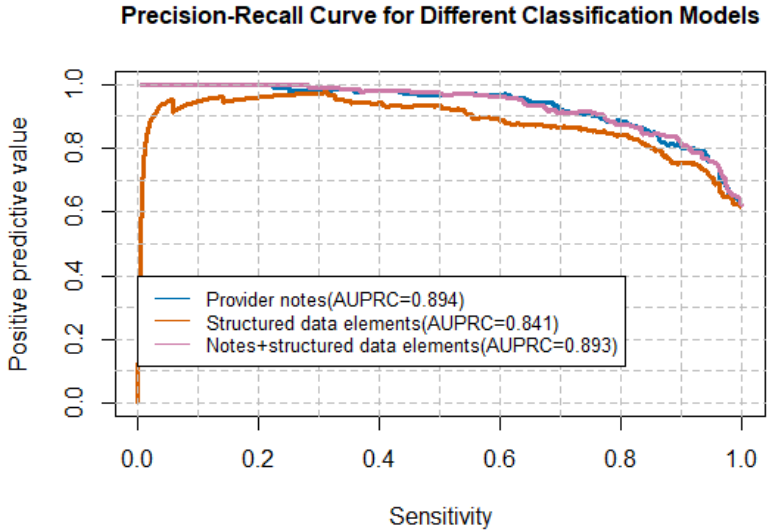

**Legend:** Precision (positive predictive value) recall (sensitivity) curve for the different classification algorithms. This illustrates the trade-off between identifying patients hospitalized due to Covid-19 (sensitivity; x-axis) from the accuracy of that capture (positive predictive value; y-axis). There is minimal difference between using just notes or notes with structured data elements. The model with only structured data elements performs notably worse in positive predictive value at the same sensitivity thresholds.

Figure2: Top words and top features

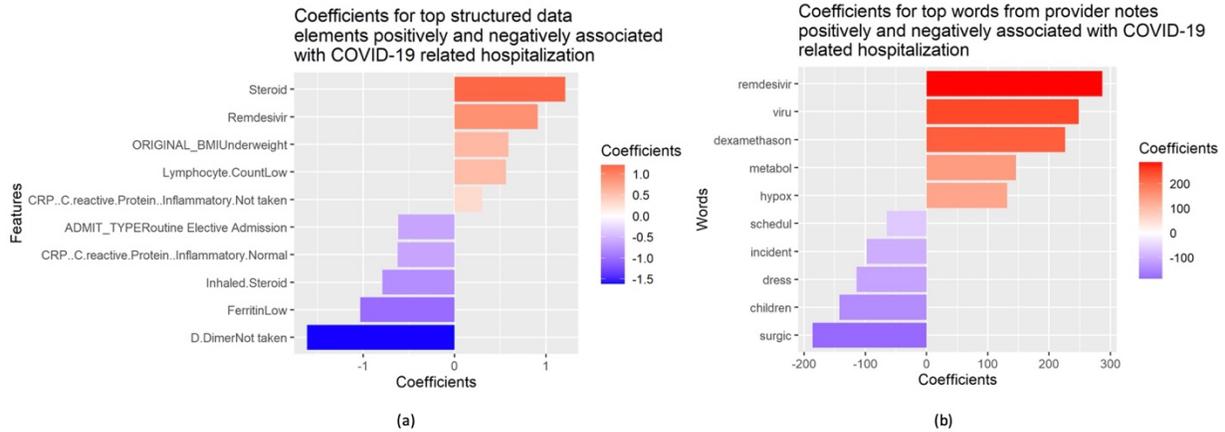

**Legend:** Top regression coefficients from the LASSO model as reflective of variable importance for the model using just structured data elements (a) and just provider notes (b). Values greater than 0 indicate the feature has positive association with hospitalization due to COVID-19, while values less than 0 indicate a feature has a negative association.

*Supplemental Figure 1:*

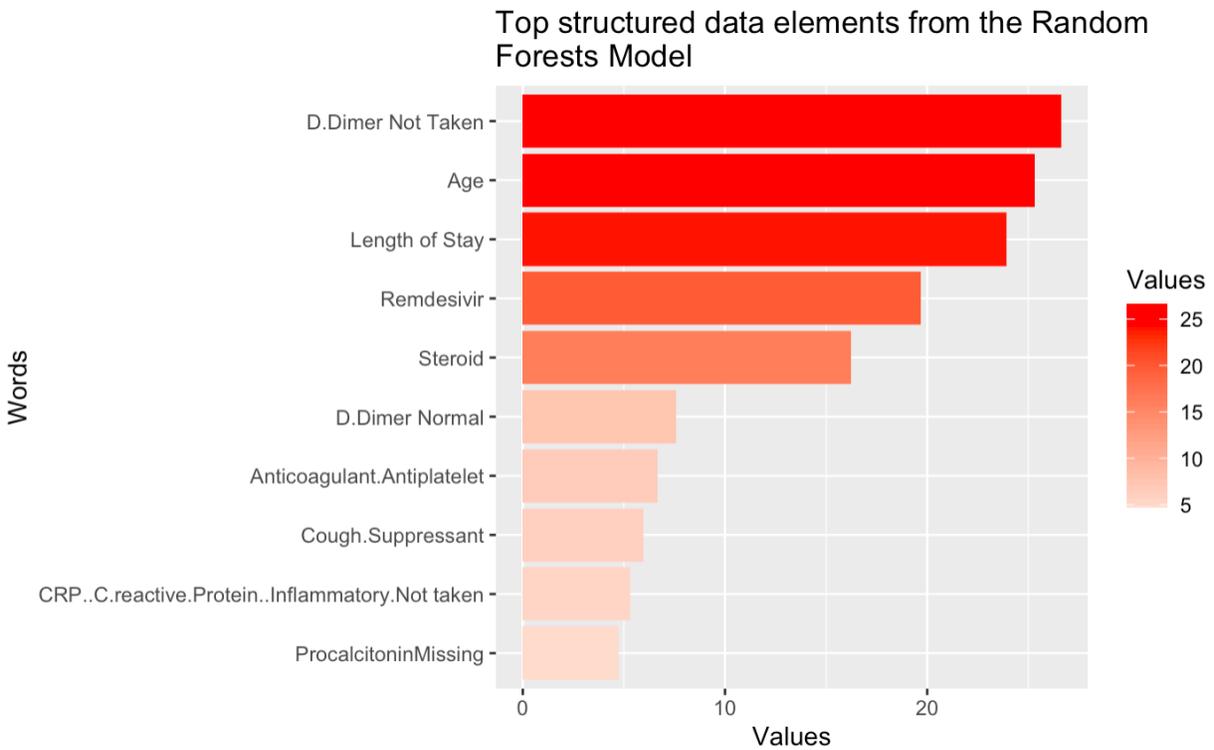

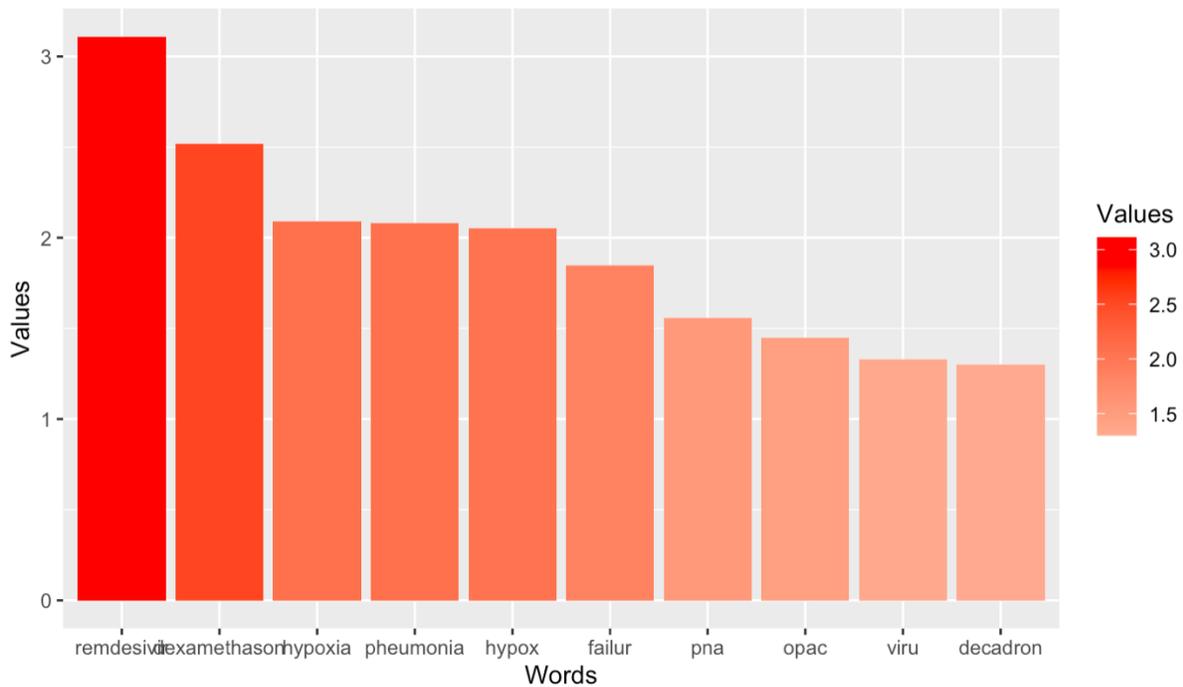

*Supplemental table 1:*

| Grouping | Variable | Definition |
| --- | --- | --- |
| Demographic | Sex | |
| | Age | |
| | Race | |
| | BMI | Category: Underweight, Normal, Obese, Overweight |
| Medical information | Patient outcome at discharge | Describe where the patient was discharged after an inpatient or emergency department encounter. |
| | Admission Type | The description of the admission type (elective surgery, etc.). |
| | Transfer to ICU | Indicates whether the patient was ever transferred to the ICU during an inpatient stay. |
| | Encounter Type | Category: Emergency Department, Emergency Department Admit to Inpatient Stay, Inpatient Hospital Stay, Observation Stay |
| | Length of Stay | Discharge Date minus Admit Date |
| | Raw payer type value | Payment Information Category: Private(commercial, managed care, medicare advantage, nc blue cross, oos blue cross), Public(medicaid pending, medicare, other |

|  |  | government, nc medicaid, nc medicaid managed care), Self-pay, Other(special programs) |
|---|---|---|
|  | Vaccine Status | COVID Vaccine Status |
| Comorbidity | Cancer | ICD-10 code: C* |
|  | Cardiovascular | ICD-10 code: I20-I59 |
|  | Hypertension | ICD-10 code: I10-I11 |
|  | Chronic Liver Disease | ICD-10 code: K70-K77 |
|  | Chronic Obstructive Pulmonary Disease | Chronic Obstructive Pulmonary Disease ICD-10 code: J44 |
|  | ASTHMA | Asthma ICD-10 code: J45 |
|  | Chronic Renal Disease | ICD-10 code: N18 |
|  | Diabetes | ICD-10 code: E08-E13 |
| Medicine | Bronchodilator | Bronchodilator |
|  | Steroid | Steroid, Corticosteroid |
|  | Anticoagulant Antiplatelet | Anticoagulant |
|  | Diuretic | Diuretic |
|  | Cough Suppressant | Cough Suppressant, Expectorant with cough suppressant |
|  | Paralytic | Paralytic, Used during intubation as a paralytic |
|  | Expectorant | Expectorant |
|  | Remdesivir | Remdesivir |
|  | Inhaled.Steroid | Inhaled Steroid |
| Lab test | Lymphocyte.Count..Absolute | The absolute number of Lymphocytes |
|  | Lymphocyte.Count | The number of Lymphocytes |
|  | CRP..C.reactive.Protein..Inflammatory | C-Reactive Protein |
|  | Ferritin | Ferritin level: High, Low, Normal, Not Taken |
|  | D.Dimer | D-Dimer level: High, Normal, Not Taken |
|  | Procalcitonin | Procalcitonin level: High, Missing, Normal |

**Supplemental Table 2:** Association between vaccine status and outcome metrics adjusted for age

| **Table 3**: Marginal association between vaccine status* and outcome | **Full Cohort (95% CI)** | **Hospitalized Due to COVID-19 (95% CI)** | **Hospitalization unrelated to COVID-19 (95% CI)** | **P-value Hospitalization due to COVID-19 vs. Hospitalization** |
|---|---|---|---|---|

| metrics, unadjusted for age Outcome | | | | unrelated to COVID-19 |
| --- | --- | --- | --- | --- |
| LOS | 0.87(0.76 0.99) | 1.02(0.86 1.20) | 0.67(0.54 0.84) | 0.001 |
| ICU | 1.12(0.75 1.65) | 1.30(0.78 2.17) | 0.87(0.44 1.71) | 0.328 |
| Mortality | 1.27(0.72 2.25) | 1.69(0.85 3.46) | 0.68(0.22 1.92) | 0.172 |

* Unvaccinated as reference group